# A Survey on Distributed Machine Learning

JOOST VERBRAEKEN, MATTHIJS WOLTING, JONATHAN KATZY, and JEROEN KLOP-
PENBURG, Delft University of Technology, Netherlands
TIM VERBELEN, imec - Ghent University, Belgium
JAN S. RELLERMEYER, Delft University of Technology, Netherlands

The demand for artificial intelligence has grown significantly over the last decade and this growth has been fueled by advances in machine learning techniques and the ability to leverage hardware acceleration. However, in order to increase the quality of predictions and render machine learning solutions feasible for more complex applications, a substantial amount of training data is required. Although small machine learning models can be trained with modest amounts of data, the input for training larger models such as neural networks grows exponentially with the number of parameters. Since the demand for processing training data has outpaced the increase in computation power of computing machinery, there is a need for distributing the machine learning workload across multiple machines, and turning the centralized into a distributed system. These distributed systems present new challenges, first and foremost the efficient parallelization of the training process and the creation of a coherent model. This article provides an extensive overview of the current state-of-the-art in the field by outlining the challenges and opportunities of distributed machine learning over conventional (centralized) machine learning, discussing the techniques used for distributed machine learning, and providing an overview of the systems that are available.

CCS Concepts: • **General and reference** → **Surveys and overviews**; • **Computing methodologies** → **Machine learning**; • **Computer systems organization** → **Distributed architectures**.

Additional Key Words and Phrases: Distributed Machine Learning, Distributed Systems

## 1 INTRODUCTION

The rapid development of new technologies in recent years has led to an unprecedented growth of data collection. Machine Learning (ML) algorithms are increasingly being used to analyze datasets and build decision making systems for which an algorithmic solution is not feasible due to the complexity of the problem. Examples include controlling self-driving cars [23], recognizing speech [8], or predicting consumer behavior [82].

In some cases, the long runtime of training the models steers solution designers towards using distributed systems for an increase of parallelization and total amount of I/O bandwidth, as the training data required for sophisticated applications can easily be in the order of terabytes [29]. In other cases, a centralized solution is not even an option when data is inherently distributed or too big to store on single machines. Examples include transaction processing in larger enterprises on data that is stored in different locations [19] or astronomical data that is too large to move and centralize [125].

In order to make these types of datasets accessible as training data for machine learning problems, algorithms have to be chosen and implemented that enable parallel computation, data distribution, and resilience to failures. A rich and diverse ecosystem of research has been conducted in this field, which we categorize and discuss in this article. In contrast to prior surveys on distributed machine learning ([120][124]) or related fields ([153][87][123][122][171][144]) we apply a wholistic view to the problem and discuss the practical aspects of state-of-the-art machine learning from a distributed systems angle.

Authors' addresses: Joost Verbraeken; Matthijs Wolting; Jonathan Katzy; Jeroen Kloppenburg, Delft University of Technology, Netherlands; Tim Verbelen, imec - Ghent University, IDLab, Department of Information Technology, Gent, Belgium; Jan S. Rellermeyer, Delft University of Technology, Faculty of Electrical Engineering, Mathematics and Computer Science, Van Mourik Broekmanweg 6, 2628XE, Delft, Netherlands.

Section 2 provides an in-depth discussion of the system challenges of machine learning and how ideas from High Performance Computing (HPC) have been adopted for acceleration and increased scalability. Section 3 describes a reference architecture for distributed machine learning covering the entire stack from algorithms to the network communication patterns that can be employed to exchange state between individual nodes. Section 4 presents the ecosystem of the most widely-used systems and libraries as well as their underlying designs. Finally, Section 5 discusses the main challenges of distributed machine learning.

## 2    MACHINE LEARNING - A HIGH PERFORMANCE COMPUTING CHALLENGE?

Recent years have seen a proliferation of machine learning technology in increasingly complex applications. While various competing approaches and algorithms have emerged, the data representations used are strikingly similar in structure. The majority of computation in machine learning workloads amount to basic transformations on vectors, matrices, or tensors–well known problems from linear algebra. The need to optimize such operations has been a highly active area of research in the high performance computing community for decades. As a result, some techniques and libraries from the HPC community (e.g., BLAS [89] or MPI [62]) have been successfully adopted and integrated into systems by the machine learning community. At the same time, the HPC community has identified machine learning to be an emerging high-value workloads and has started to apply HPC methodology to them. Coates et al. [38] were able to train a 1 billion parameter network on their Commodity Off-The-Shelf High Performance Computing (COTS HPC) system in just three days. You et al. [166] optimized the training of a neural network on Intel's Knights Landing, a chip designed for HPC applications. Kurth et al. [84] demonstrated how deep learning problems like extracting weather patterns can be optimized and scaled efficiently on large parallel HPC systems. Yan et al. [163] have addressed the challenge of scheduling deep neural network applications on cloud computing infrastructure by modeling the workload demand with techniques like lightweight profiling that are borrowed from HPC. Li et al. [91] investigated the resilience characteristics of deep neural networks with regard to hardware errors when running on accelerators, which are frequently deployed in major HPC systems.

Like for other large-scale computational challenges, there are two fundamentally different and complementary ways of accelerating workloads: adding more resources to a single machine (vertical scaling or scaling-up) and adding more nodes to the system (horizontal scaling or scaling-out).

### 2.1    Scaling Up

Among the scale-up solutions, adding programmable GPUs is the most common method and various systematic efforts have shown the benefits of doing so (e.g., [126], [18], [78]). GPUs feature a high number of hardware threads. For example, the Nvidia Titan V and Nvidia Tesla V100 have a total of 5120 cores which makes them approximately 47x faster for deep learning than a regular server CPU (namely an Intel Xeon E5-2690v4) [108]. Originally the applications of GPUs for machine learning were limited because GPUs used a pure SIMD (Single Instruction, Multiple Data) [51] model that did not allow the cores to execute a different branch of the code; all threads had to perform the exact same program. Over the years GPUs have shifted to more flexible architectures where the overhead of branch divergence is reduced, but diverging branches is still inefficient [66]. The proliferation of GPGPUs (General-Purpose GPUs, i.e. GPUs that can execute arbitrary code) has lead the vendors to design custom products that can be added to conventional machines as accelerators and no longer fulfill any role in the graphics subsystem of the machine. For example, the Nvidia Tesla GPU series is meant for highly parallel computing and designed for deployment in supercomputers and clusters. When a sufficient degree of parallelism is offered by the workload, GPUs can significantly accelerate machine learning algorithms. For example, Meuth [101] reported a speed-up up to 200x over conventional CPUs for an image recognition algorithm using a pretrained multilayer perceptron (MLP).

An alternative to generic GPUs for acceleration is the use of Application Specific Integrated Circuits (ASICs) which implement specialized functions through a highly optimized design. In recent times, the demand for such chips has risen significantly [100]. When applied to e.g. Bitcoin mining, ASICs have a significant competitive advantage over GPUs and CPUs due to their high performance and power efficiency [145]. Since matrix multiplications play a prominent role in many machine learning algorithms, these workloads are highly amenable to acceleration through ASICS. Google applied this concept in their Tensor Processing Unit (TPU) [129], which, as the name suggests, is an ASIC that specializes in calculations on tensors ($n$-dimensional arrays), and is designed to accelerate their Tensorflow [1][2] framework, a popular building block for machine learning models. The most important component of the TPU is its Matrix Multiply unit based on a systolic array. TPUs use a MIMD (Multiple Instructions, Multiple Data) [51] architecture which, unlike GPUs, allows them to execute diverging branches efficiently. TPUs are attached to the server system through the PCI Express bus. This provides them with a direct connection with the CPU which allows for a high aggregated bandwidth of 63GB/s (PCI-e5x16). Multiple TPUs can be used in a data center and the individual units can collaborate in a distributed setting. The benefit of the TPU over regular CPU/GPU setups is not only its increased processing power but also its power efficiency, which is important in large-scale applications due to the cost of energy and the limited availability in large-scale data centers. When running benchmarks, Jouppi et al. [80] found that the performance per watt of a TPU can approach 200x that of a traditional system. Further benchmarking by Sato et al. [129] indicated that the total processing power of a TPU or GPU can be up to 70x higher than a CPU for a typical neural network, with performance improvements varying from 3.5x - 71x depending on the task at hand.

Chen et al. [32] developed DianNao, a hardware accelerator for large-scale neural networks with a small area footprint. Their design introduces a Neuro-Functional Unit (NFU) in a pipeline that multiplies all inputs, adds the results, and, in a staggered manner after all additions have been performed, optionally applies an activation function like a sigmoid function. The experimental evaluation using the different layers of several large neural network structures ([48], [90], [134], [133], [70]) shows a performance speedup of three orders of magnitude and an energy reduction of more than 20x compared to using a general-purpose 128bit 2GHz SIMD CPU.

Hinton et al. [70] address the challenge that accessing the weights of neurons from DRAM is a costly operation and can dominate the energy profile of processing. Leveraging a *deep compression* technique, they are able to put the weights into SRAM and accelerate the resulting sparse matrix-vector multiplications through efficient weight sharing. The result is a 2.9 times higher throughput and a 19x improved energy efficiency compared to DianNao.

Even general purpose CPUs have increased the availability and width of vector instructions in recent product generations in order to accelerate the processing of computationally intensive problems like machine learning algorithms. These instructions are vector instructions part of the AVX-512 family [127] with enhanced word-variable precision and support for single precision floating-point operations. In addition to the mainstream players, there are also more specialized designs available such as the Epiphany [111]. This special-purpose CPU is designed with a MIMD architecture that uses an array of processors, each of which accessing the same memory, to speed up execution of floating point operations. This is faster than giving every processor its own memory because communicating between processors is expensive. The newest chip of the major manufacturer Adapteva is the Epiphany V, which contains 1024 cores on a single chip [110]. Although Adapteva has not published power consumption specifications of the Epiphany V yet, it has released numbers suggesting a power usage of only 2 Watt [4].

## 2.2 Scaling Out

While there are many different strategies to increase the processing power of a single machine for large-scale machine learning, there are reasons to prefer a scale-out design or combine the two approaches, as often seen in HPC. The first reason is the generally lower equipment cost, both in terms of initial investment and maintenance. The second reason is the resilience against failures because, when a single processor fails within an HPC application, the system can still continue operating by initiating a partial recovery (e.g., based on communication-driven checkpointing [46] or partial re-computation [169]). The third reason is the increase in aggregate I/O bandwidth compared to a single machine [49]. Training ML models is a highly data-intensive task and the ingestion of data can become a serious performance bottleneck [67]. Since every node has a dedicated I/O subsystem, scaling out is an effective technique for reducing the impact of I/O on the workload performance by effectively parallelizing the reads and writes over multiple machines. A major challenge of scaling-out is that not all ML algorithms lend themselves to a distributed computing model which can thus only be used for algorithms that can achieve a high degree of parallelism.

## 2.3 Discussion

The lines between traditional supercomputers, grids, and the cloud are increasingly getting blurred when it comes to the best execution environment for demanding workloads like machine learning. For instance, GPUs and accelerators are now more common in major cloud datacenters [135][136]. As a result, parallelization of the machine learning workload has become paramount to achieving acceptable performance at large scale. When transitioning from a centralized solution to a distributed system, however, the typical challenges of distributed computing in the form of performance, scalability, failure resilience, or security apply [40]. The following section presents a systematic discussion of the different aspects of distributed machine learning and develops a reference architecture by which all existing systems can be categorized.

## 3 A REFERENCE ARCHITECTURE FOR DISTRIBUTED MACHINE LEARNING

Designing a generic system that enables an efficient distribution of regular Machine Learning is challenging since every algorithm has a distinct communication pattern [78][106][128][146][150][152]. Despite various different concepts and implementations for distributed machine learning, we have identified a common architectural framework that covers the entire design space. Every section discusses a particular area where designers of machine learning solutions need to make a decision.

In general, the problem of machine learning can be separated into the training and the prediction phase (Figure 1).

The *Training phase* involves training a machine learning model by feeding it a large body of training data and updating it using an ML algorithm. An overview of applicable and commonly-used algorithms is given in Section 3.1. Aside from choosing a suitable algorithm for a given problem, we also need to find an optimal set of hyperparameters for the chosen algorithm, which is described in Section 3.2. The final outcome of the training phase is a *Trained Model*, which can then be deployed. The *Prediction phase* is used for deploying the trained model in practice. The trained model accepts new data as input and produces a prediction as output. While the training phase of the model is typically computationally intensive and requires the availability of large data sets, the inference can be performed with less computing power.

The training phase and prediction phase are not mutually exclusive. Incremental learning combines the training phase and inference phase and continuously trains the model by using new data from the prediction phase.

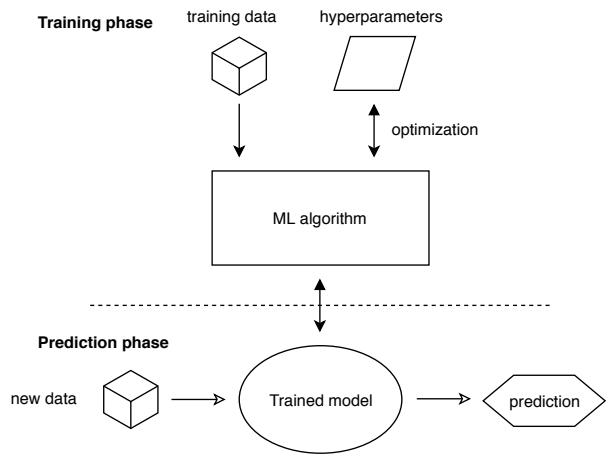

Fig. 1. General Overview of Machine Learning. During the training phase a ML model is optimized using training data and by tuning hyper parameters. Then the trained model is deployed to provide predictions for new data fed into the system.

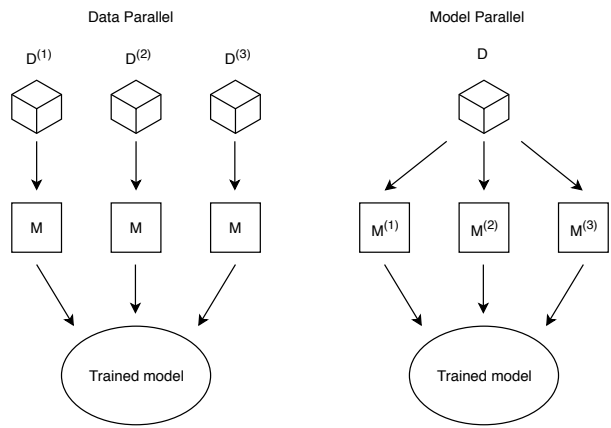

Fig. 2. Parallelism in Distributed Machine Learning. Data parallelism trains multiple instances of the same model on different subsets of the training dataset, while model parallelism distributes parallel paths of a single model to multiple nodes.

When it comes to distribution, there are two fundamentally different ways of partitioning the problem across all machines, parallelizing the data or the model [120] (Figure 2). These two methods can also be applied simultaneously [162].

In the *Data-Parallel* approach, the data is partitioned as many times as there are worker nodes in the system and all worker nodes subsequently apply the same algorithm to different data sets. The same model is available to all worker nodes (either through centralization, or through replication) so that a single coherent output emerges naturally. The technique can be used with every ML algorithm with an independent and identically distribution (i.i.d.) assumption over the data samples (i.e. most ML algorithms [162]). In the *Model-Parallel* approach, exact copies of the entire data sets are processed by the worker nodes which operate on different parts of the model. The model is therefore the aggregate of all model parts. The model-parallel approach cannot automatically be applied to every machine learning algorithms because the model parameters generally cannot be split up. One option is to train different instances of the same or similar model, and aggregate the outputs of all trained models using methodologies like ensembling (Section 3.3).

The final architectural decision is the *topology* of the distributed machine learning system. The different nodes that form the distributed system need to be connected through a specific architectural pattern in order to fulfill a common task. However, the choice of pattern has implications on the role that a node can play, the degree of communication between nodes, and the failure resilience of the whole deployment. A discussion of commonly used topologies is presented in Section 3.4.

In practice, the three layers of architecture (machine learning, parallelism, topology) are not independent. The combining factor is their impact on the amount of communication required to train the model, which is discussed in Section 3.5.

## 3.1 Machine Learning Algorithms

Machine Learning algorithms learn to make decisions or predictions based on data. We categorize current ML algorithms based on the following three characteristics:

- **Feedback**, the type of feedback that is given to the algorithm while learning
- **Purpose**, the desired end result of the algorithm
- **Method**, the nature of model evolution that occurs when given feedback

### 3.1.1 Feedback.

To train an algorithm, it requires feedback so that it can gradually improve the quality of the model. There are several different types of feedback [165]:

- **Supervised learning** uses training data that consists of input objects (usually vectors) and the corresponding desired output values. Supervised learning algorithms attempt to find a function that maps the input data to the desired output. Then, this function can be applied to new input data to predict the output. One of the goals is to minimize both the bias and variance error of the predicted results. The bias error is caused by simplifying assumptions made by the learning algorithm in order to facilitate learning the target function. However, methods with high bias have lower predictive performance on problems that do not fully satisfy the assumptions. For example, a linear model will not be able to give accurate predictions if the underlying data has a non-linear behavior. The variance captures how much the results of the ML algorithm change for a different train set. A high variance means that the algorithm is modeling the specifics of the training data, without finding the underlying (hidden) mapping between the inputs and the outputs. Unfortunately, eliminating both the bias and the variance is typically impossible, a phenomenon known as the bias-variance trade-off [54]. The more complex the model, the more training data is required to train the algorithm to gain an accurate prediction from the model. For example, when the dimensionality of the data is high, the output may depend on a convoluted combination of input factors which requires a high number of data samples to detect the relations between these dimensions.
- **Unsupervised learning** uses training data that consists of input objects (usually vectors) without output values. Unsupervised learning algorithms aim at finding a function that describes the structure of the data and group the unsorted input data. Because the input data is unlabeled, it lacks a clear output accuracy metric. The most common use case of unsupervised learning is to cluster data together based on similarities and hidden patterns. Unsupervised learning is also used for problems like dimensionality reduction where the key features of data are extracted. In this case the feedback is generated using a similarity metric.
- **Semi-supervised learning** uses a (generally small) amount of labeled data, supplemented by a comparatively large amount of unlabeled data. Clustering can be used to extrapolate known labels onto unlabeled data points. This is done under the assumption that similar data points share the same label.

- **Reinforcement learning** is used to train an agent that has to take actions in an environment based on its observations. Feedback relies on a reward or cost function that evaluates the states of the system. The biggest challenge here is the credit assignment problem, or how to determine which actions actually lead to higher reward in the long run. Bagnell and Ng [13] showed that a local reward system is beneficial for the scalability of the learning problem since global schemes require samples that scale roughly linearly with the number of participating nodes.

### 3.1.2 Purpose.

Machine learning algorithms can be used for a wide variety of purposes, such as classifying an image or predicting the probability of an event. They are often used for the following tasks [85]:

- **Anomaly detection** is used to identify data samples that differ significantly from the majority of the data. These anomalies, which are also called outliers, are used in a wide range of applications including video surveillance, fraud detection in credit card transactions or health monitoring with on-body sensors.
- **Classification** is the problem of categorizing unknown data points into categories seen during training. This is an inherently supervised process; the unsupervised equivalent of classification is clustering.
- **Clustering** groups data points that are similar according to a given metric. Small data sets can be clustered by manually labeling every instance, but for larger datasets that might be infeasible, which justifies the need for automatic labeling the instances (namely clustering)
- **Dimensionality reduction** is the problem of reducing the number of variables in the input data. This can either be achieved by selecting only relevant variables (Feature selection), or by creating new variables that represent multiple others (Feature extraction).
- **Representation learning** attempts to find proper representations of input data for, e.g., feature detection, classification, clustering, encoding, or matrix factorization. This often also implies a dimensionality reduction.
- **Regression** is the problem of estimating how a so-called *dependent* variable changes in value when other variables change with a certain amount.

### 3.1.3 Method.

Every effective ML algorithm needs a method that forces the algorithm to improve itself based on new input data so that it can improve its accuracy. We identify five different groups of ML methods that distinguish themselves through the way the algorithm learns:

- **Evolutionary algorithms (EAs) [57]** (and specifically **Genetic algorithms**) learn iteratively based on evolution. The model that actually solves the problem is represented by a set of properties, called its *genotype*. The performance of the model is measured using a score, calculated using a *fitness function*. After calculating the fitness score of all generated models, the next iteration creates new genotypes based on mutation and crossover of models that produce more accurate estimates. Genetic algorithms can be used to create other algorithms, such as neural networks, belief networks, decision trees, and rule sets.
- **Stochastic gradient descent (SGD) based algorithms** minimize a loss function defined on the outputs of the model by adapting the model's parameters in the direction of the negative gradient (the multi-variable derivative of a function)). The gradient descent is called stochastic as the gradient is calculated from a randomly sampled subset of the training data. The loss function is typically a proxy for the actual error to be minimized, for example the mean squared error between the model outputs and desired outputs in the case of a regression problem, or the negative log likelihood of the ground truth class according to the model in the case of classification. The typical training procedure then becomes:

(1) Present a batch of randomly sampled training data.
(2) Calculate the loss function of the model output and the desired output.
(3) Calculate the gradient with respect to the model parameters.
(4) Adjust the model parameters in the direction of the negative gradient, multiplied by a chosen learning rate.
(5) Repeat

SGD is the most commonly used training method for a variety of ML models.

– **Support vector machines (SVMs)** map data points to high dimensional vectors for classification and clustering purposes. For data points in a p-dimensional space, a (p-1)-dimensional hyperplane can be used as a classifier. A reasonable choice would be the hyperplane that properly separates the data points in two groups based on their labels by the largest possible margin. Sometimes special transformation equations (called kernels) are used to transform all data points to a different representation, in which it is easier to find such a hyperplane.

– **Perceptrons [105]** are binary classifiers that label input vectors as 'active' or 'inactive'. A Perceptron assign a weight to all inputs and then sums over the products of these weights and their input. The outcome of this is compared to a threshold in order to determine the label. Perceptron-based algorithms commonly use the entire batch of training data in their attempt to find a solution that is optimal for the whole set. They are binary, and therefore primarily used for binary classification.

– **Artificial neural networks (ANNs)** are perceptron-based systems that consist of multiple layers: an input layer, one or more hidden layers and an output layer. Each layer consists of nodes connected to the previous and next layers through edges with associated weights (usually called synapses). Unlike regular perceptrons, these nodes usually apply an activation function on the output to introduce non-linearities.

The model is defined by the state of the entire network, and can be changed by altering (1) the weights of the synapses, (2) the layout of the network, or (3) the activation function of nodes.

Because neural networks require a large number of nodes, the understandability of a neural network's *thought process* is lower compared to e.g. decision trees.

Neural networks are extensively studied because of their ability to analyze enormous sets of data. They can be categorized into several subgroups based on network layout:

  ∗ **Deep neural networks (DNNs)**, are artificial neural networks that have many hidden layers. This allows the neural network to learn hierarchical feature abstractions of the data, with increasing abstraction the deeper you go in the network.

  ∗ **Convolutional neural networks (CNNs / ConvNets)** are deep, feed-forward neural networks that use convolution layers with nodes connected to only a few nodes in the previous layer. These values are then pooled using pooling layers. It can be seen as a way of recognizing abstract features in the data. The convolution makes the network consider only local data. This makes the represented algorithms spatially invariant, which is why they are sometimes called Space Invariant Artificial Neural Networks (SIANN). Chaining multiple of these convolution and pooling layers together can make the network capable of recognizing complicated constructs in big datasets. Examples of this are cats in images or the contextual meaning of a sentence in a paragraph.

  ∗ **Recurrent neural networks (RNNs)** keep track of a temporal state in addition to weights, which means that previous inputs of the network influence its current decisions.

Recurrent synapses give the network a *memory*. This can help with discovering temporal patterns in data. Blocks of nodes in recurrent network operate as cells with distinct memories, and can store information for an arbitrarily long timespan.

* **Hopfield networks** are a type of non-reflexive, symmetric recurrent neural network that have an *energy* related to every state of the network as a whole. They are guaranteed to converge on a local minimum after some number of network updates.

* **Self-organizing maps (SOMs) / self-organizing feature maps (SOFMs)** are neural networks that learn through unsupervised *competitive learning*, in which nodes compete for access to specific inputs. This causes the nodes to become highly specialized, which reduces redundancy. The iterations effectively move the map closer to the training data, which is the reason for its name. Some subtypes include the Time Adaptive Self-Organizing Map (TASOM, automatically adjust the learning rate and neighborhood size of each neuron independently), Binary Tree TASOM (BTASOM, tree of TASOM networks) and Growing Self-Organizing map (GSOM, identify a suitable map size in the SOM by starting with a minimal set of nodes and growing the map by heuristically adding new nodes at the periphery).

* **Stochastic neural networks** make use of stochastic transfer functions or stochastic weights, which allows them to escape the local minima that impede the convergence to a global minimum of normal neural networks. An example is a Boltzmann machine where each neuron output is represented as a binary value and the likelihood of the neuron firing depends on the network of other neurons.

* **Auto-encoders** are a type of neural network that are trained specifically to encode and decode data. Since auto-encoders are trained to perform decoding separately from encoding, the encoded version of the data is a form of dimensionality reduction of the data.

* **Generative Adversarial Networks (GAN)** are generative models that are trained using a minimax game between a generator and discriminator network [58]. The goal is to train a neural network to generate data from a training set distribution. To achieve this, a discriminator neural network is trained at the same time to learn to discriminate between real dataset samples and generated samples by the generator. The discriminator is trained to minimize the classification errors, whereas the generator is trained to maximize the classification errors, in effect generating data that is indistinguishable from the real data.

* **Rule-based machine learning (RBML) algorithms** [157] use a set of rules that each represent a small part of the problem. These rules usually express a condition, as well as a value for when that condition is met. Because of the clear if-then relation, rules lend themselves to simple interpretation compared to more abstract types of ML algorithms, such as neural networks.

  - **Association Rule Learning** is a *rule-based machine learning* method that focuses on finding relations between different variables in datasets. Example relatedness metrics are *Support* (how often variables appear together), *Confidence* (how often a causal rule is true) and *Collective Strength* (inverse likelihood of the current data distribution if a given rule does not exist).

  - **Decision trees**, sometimes called CART trees after Classification And Regression Trees, use rule-based machine learning to create a set of rules and decision branches. Traversing the tree involves applying the rules at each step until a leaf of the tree is reached. This leaf represents the decision or classification for that input.

* **Topic Models (TM)** [21] are statistical models for finding and mapping semantic structures in large and unstructured collections of data, most often applied on text data.

  – **Latent Dirichlet Allocation** [22] constructs a mapping between documents and a proba-
    bilistic set of topics, using the assumption that documents have few different topics and
    that those topics use few different words. It is used to learn what unstructured documents
    are about based on a few keywords.
  – **Latent semantic analysis (LSA) / latent semantic indexing (LSI)** creates a big matrix
    of documents and topics in an attempt to classify documents or to find relations between
    topics. LSA/LSI assumes a Gaussian distribution for topics and documents. LSA/LSI does
    not have a way of dealing with words that have multiple meanings.
  – **Naive Bayes classifiers** are relatively simple probabilistic classifiers that assume different
    features to be independent. They can be trained quickly using supervised learning, but are
    less accurate than more complicated approaches.
  – **Probabilistic latent semantic analysis (PLSA) / probabilistic latent semantic index-
    ing (PLSI)** is the same as LSA/LSI, except that PLSA/PLSI assumes a Poisson distribution
    for topics and documents instead of the Gaussian distribution that is assumed by LSA/LSI.
    The reason is that a Poisson distribution appears to model the real world better [72]. Some
    subtypes include Multinomial Asymmetric Hierarchical Analysis (MASHA), Hierarchical
    Probabilistic Latent Semantic Analysis (HPLSA), and Latent Dirichlet Allocation (LDA).
- **Matrix Factorization** algorithms can be applied for identifying latent factors or find missing
  values in matrix-structured data. For example many recommender systems are based on
  matrix factorization of the *User-Item Rating Matrix* to find new items users might be interested
  in given their rating on other items [83]. Similarly factorizing a *Drug compound-Target Protein
  Matrix* is used for new drug discovery [63]. As this problem scales with $O(F^3)$ with $F$ the
  dimensionality of the features, recent research focuses on scaling these methods to larger
  feature dimensions [143].

## 3.2 Hyperparameter Optimization

The performance of many of the algorithms presented in the previous sections are largely impacted
by the choice of a multitude of algorithm hyperparameters. For example, in stochastic gradient
descent, one has to choose the batch size, the learning rate, the initialization of the model, etc.
Often, the optimal values of these hyperparameters are different for each problem domain, ML
model, and dataset.

There are several algorithms that can be used to automatically optimize the parameters of the
machine learning algorithms and that can be re-used across different Machine Learning algorithm
families.

These include:

- First-order algorithms that use at least one first-derivative of the function that maps the
  parameter value to the accuracy of the ML algorithm using that parameter. Examples are
  stochastic gradient descent (SGD) [24], stochastic dual coordinate ascent [137], or conjugate
  gradient methods [69][42]
- Second-order techniques that use any second-derivative of the function that maps the param-
  eter value to the accuracy of the ML algorithm using that parameter. Examples are Newton's
  method [121] (which requires computing the Hessian matrix, and is therefore generally
  infeasible), Quasi-Newton methods [28] (which approximate Newton's method by updating
  the Hessian by analyzing successive gradient vectors instead of recomputing the Hessian in
  every iteration), or L-BFGS [95].

- Coordinate descent [159] (also called *coordinate-wise minimization*), which minimizes at each iteration a single variable while keeping all other variables at their value of the current iteration.
- The Markov-Chain Monte-Carlo [26], which works by successively guessing new parameters randomly drawn from a normal multivariate solution centered on the old parameters and using these new parameters with a chance dependent on the likelihood of the old and the new parameters.
- A naive but often used strategy is grid search, which exhaustively runs to a grid of potential values of each hyperparameter [88].
- Random search uses randomly chosen trials for sampling hyperparameter values, which often yields better results in terms of efficiency compared to grid search, finding better parameter values for the same compute budget [17].
- Bayesian hyperparameter optimization techniques use the Bayesian framework to iteratively sample hyperparameter values [147]. These model each trial as a sample form a Gaussian process (GP), and use the GP to choose the most informative samples in the next trial.

### 3.3 Combining Multiple Algorithms: Ensemble Methods

For some applications, a single model is not accurate enough to solve the problem. To alleviate this issue, multiple models can be combined in so-called *Ensemble Learning*. For example when machine learning algorithms are performed on inherently distributed data sources and centralization is thus not an option, the setup requires training to happen in two separate stages: first in the local sites where the data is stored and second in the global site that aggregates the over the individual results of the first stage [77]. This aggregation can be achieved by applying ensemble methods in the global site.

Various different ways exist to perform ensembling, such as [50]:

- **Bagging** is the process of building multiple classifiers and combining them into one.
- **Boosting** is the process of training new models with the data that is misclassified by the previous models.
- **Bucketing** is the process of training many different models and eventually selecting the one that has the best performance.
- **Random Forests** [25] use multiple decision trees and averaging the prediction made by the individual trees as to increase the overall accuracy. Different trees are given the same 'voting power'.
- **Stacking** is when multiple classifiers are trained on the dataset, and one new classifier uses the output of the other classifiers as input in an attempt to reduce the variance.
- **Learning Classifier Systems (LCSs)** is a modular system of learning approaches. An LCS iterates over data points from the dataset, completing the entire learning process in each iteration. The main idea is that an LCS has a limited number of rules. A Genetic Algorithm forces suboptimal rules out of the rule set. There are many different attributes that can drastically change the performance of an LCS depending on the dataset, including the Michigan-style vs Pittsburgh-style architecture [114], supervised vs reinforcement learning [81], incremental vs batch learning [37], online vs offline training, strength-based vs accuracy-based [158] and complete mapping vs best mapping.

### 3.4 Topologies

Another consideration for the design of a Distributed Machine Learning deployment is the structure in which the computers within the cluster are organized. A deciding factor for the topology is the degree of distribution that the system is designed to implement. Figure 3 shows four possible topologies, in accordance with the general taxonomy of distributed communication networks by Baran [15].

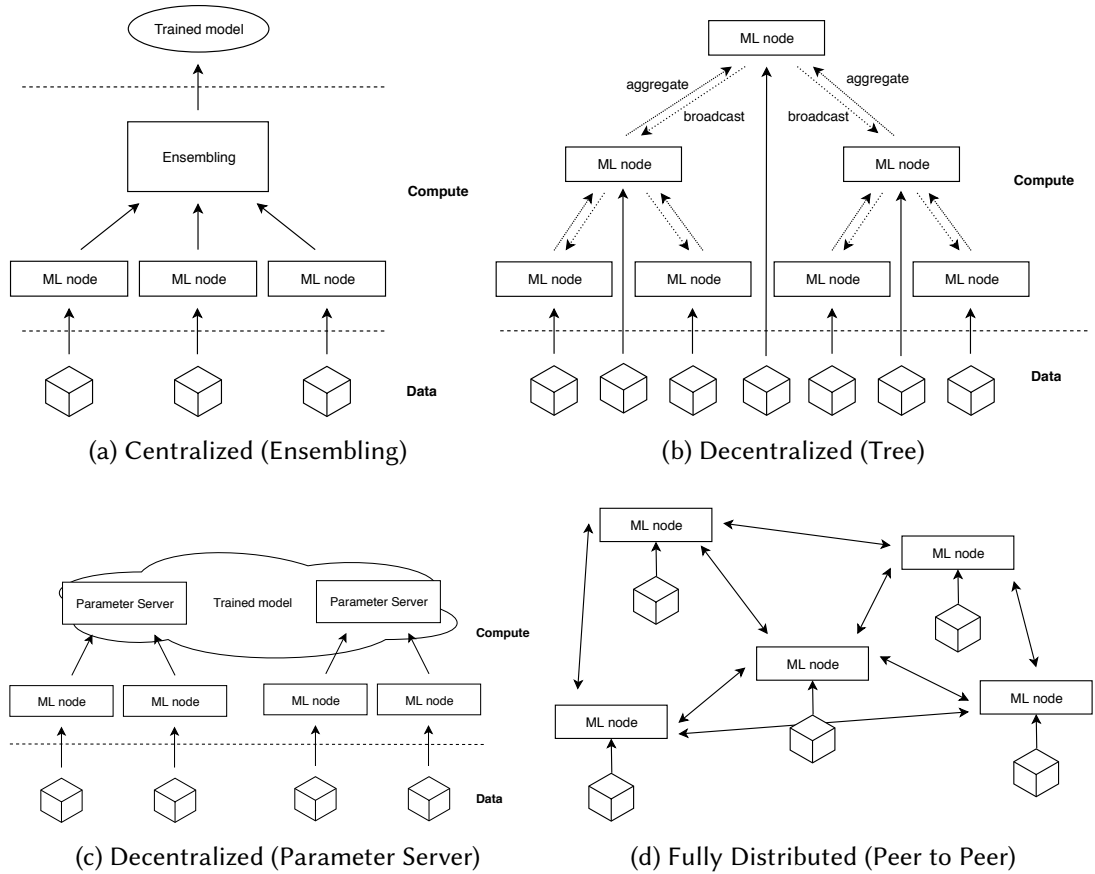

(a) Centralized (Ensembling)  (b) Decentralized (Tree)

(c) Decentralized (Parameter Server)  (d) Fully Distributed (Peer to Peer)

Fig. 3. Distributed Machine Learning Topologies, based on the degree of distribution.

*Centralized systems* (Figure 3a) employ a strictly hierarchical approach to aggregation, which happens in a single central location. *Decentralized systems* allow for intermediate aggregation, either with a replicated model that is consistently updated when the aggregate is broadcast to all nodes such as in tree topologies (Figure 3b) or with a partitioned model that is sharded over multiple parameter servers (Figure 3c). *Fully distributed systems* (Figure 3d) consists of a network of independent nodes that ensemble the solution together and where no specific roles are assigned to certain nodes.

There are several distinct topologies that have become popular choices for distributed machine learning clusters:

- **Trees** Tree-like topologies have the advantage that they are easy to scale and manage, as each node only has to communicate with its parent and child nodes. For example in the AllReduce [5] paradigm, nodes in a tree accumulate their local gradients with those from their children and pass this sum to their parent node in order to calculate a global gradient.
- **Rings** In situations where the communication system does not provide efficient support for broadcast or where communication overhead needs to be kept to a minimum, ring topologies for AllReduce patterns simplify the structure by only requiring neighbor nodes to synchronize through messages. This is, e.g., commonly used between multiple GPUs on the same machine [76].

- **Parameter Server** The Parameter Server paradigm (PS) [156] uses a decentralized set of workers with a centralized set of masters that maintain the shared state. All model parameters are stored in a shard on each Parameter Server, from which all clients read and write as a key-value store. An advantage is that all model parameters (within a shard) are in a global shared memory, which makes it easy to inspect the model. A disadvantage of the topology is that the Parameter Servers can form a bottleneck, because they are handling all communication. To partially alleviate this issue, the techniques for bridging computation and communication mentioned in Section 3.5.2 are used.

- **Peer to Peer** In contrast to centralized state, in the fully distributed model, every node has its own copy of the parameters, and the workers communicate directly with each other. This has the advantage of typically higher scalability than a centralized model and the elimination of single points of failure in the system [52]. An example implementation of this model is a peer-to-peer network in which nodes broadcast updates to all other nodes to form a data-parallel processing framework. Since full broadcast is typically prohibitive due to the volume of communication, Sufficient Factor Broadcasting (SFB) [94]) has been proposed to reduce the communication overhead. The parameter matrix in SFB is decomposed into so-called sufficient factors, i.e. 2 vectors that are sufficient to reconstruct the update matrix. SFB only broadcasts these sufficient factors and lets the workers reconstruct the updates. Other models limit the degree of communication to less frequent synchronization points while allowing the individual models to temporarily diverge. Gossip Learning [139] is built around the idea that models are mobile and perform independent random walks through the peer-to-peer network. Since this forms a data- and model-parallel processing framework, the models evolve differently and need to be combined through ensembling. In Gossip Learning, this happens continuously on the nodes by combining the current model with a limited cache of previous visitors.

## 3.5 Communication

As previously discussed, the need for more sophisticated machine learning-based setups quickly outgrows the capabilities of a single machine. There are several ways to partition the data and/or the program and to distribute these evenly across all machines. The choice of distribution, however, has direct implications on the amount of communication required to train the model.

*3.5.1 Computation Time vs. Communication vs. Accuracy.* When Distributed Machine Learning is used, one aims for the best accuracy at the lowest computation and communication cost. However, for complex ML problems, the accuracy usually increases with processing more training data, and sometimes by increasing the ML model size, hence increasing the computation cost. Parallelizing the learning can reduce computation time, as long as the communication costs are not becoming dominant. This can become a problem if the model being trained is not sufficiently large in comparison to the data. If the data is already distributed (e.g., cloud-native data), there is no alternative to either moving the data or the computation.

Splitting up the dataset across different machines and training a separate model on a separate part of the dataset avoids communication, but this reduces the accuracy of the individual models trained on each machine. By ensembling all these models, the overall accuracy can be improved, However, the computation time is typically not much lower, since the individual models still have to take the same number of model update steps in order to converge.

By already synchronizing the different models during training (e.g., by combining the calculated gradients on all machines in case of gradient descent), the computation time can be reduced by converging faster to a local optimum. This, however, leads to an increase of communication cost as the model size increases.

Therefore, practical deployments require seeking the amount of communication needed to achieve the desired accuracy within an acceptable computation time.

*3.5.2 Bridging Computation and Communication.* To schedule and balance the workload, there are three concerns that have to be taken into account [162]:

- Identifying which tasks can be executed in parallel
- Deciding the task execution order
- Ensuring a balanced load distribution across the available machines

After deciding on these three issues, the information between nodes should be communicated as efficiently as possible. There are several techniques that enable the interleaving of parallel computation and inter-worker communication. These techniques trade off fast / correct model convergence (at the top of the list found below) with faster / fresher updates (at the bottom of the list found below).

- **Bulk Synchronous Parallel (BSP)** is the simplest model in which programs ensure consistency by synchronizing between each computation and communication phase [162]. An example of program following the BSP bridging model is MapReduce.
  An advantage is that serializable BSP ML programs are guaranteed to output a correct solution. A disadvantage is that finished workers must wait at every synchronization barrier until all other workers are finished, which results in overhead in the event of some workers progressing slower than others [34].
- **Stale Synchronous Parallel (SSP)** relaxes the synchronization overhead by allowing the faster workers to move ahead for a certain number of iterations. If this number is exceeded, all workers are paused. Workers operate on cached versions of the data and only commit changes at the end of a task cycle, which can cause other workers to operate on stale data. The main advantage of SSP is that it still enjoys strong model convergence guarantees. A disadvantage however, is that when the staleness becomes too high (e.g. when a significant number of machines slows down), the convergence rates quickly deteriorate. The algorithm can be compared to Conits [167] used in distributed systems, because it specifies the data on which the workers are working and consistency is to be measured.
- **Approximate Synchronous Parallel (ASP)** limits how inaccurate a parameter can be. This contrasts with SSP, which limits how stale a parameter can be. An advantage is that, whenever an aggregated update is insignificant, the server can delay synchronization indefinitely. A disadvantage is that it can be hard to choose the parameter that defines which update are significant and which are not [73].
- **Barrierless Asynchronous Parallel [65] / Total Asynchronous Parallel [73] (BAP / TAP)** lets worker machines communicate in parallel without waiting for each other. The advantage is that it usually obtains the highest possible speedup. A disadvantage is that the model can converge slowly or even develop incorrectly because, unlike BSP and SSP, the error grows with the delay [65].

*3.5.3 Communication Strategies.* Communication is an important contributor to defining the performance and scalability of distributed processing [27]. Several communication management strategies [162] are used to spread and reduce the amount of data exchanged between machines:

- To prevent bursts of communication over the network (e.g. after a mapper is finished), continuous communication is used, such as in the state-of-the-art implementation Bösen [156].
- Neural networks are composed out of layers, the training of which (using the back-propagation gradient descent algorithm) is highly sequential. Because the top layers of neural networks contain the most parameters while accounting only for a small part of the total computation, Wait-free Backpropagation (WFBP) [172] was proposed. WFBP exploits the neural network structure by already sending out the parameter updates of the top layers while still computing the updates for the lower layers, hence hiding most of the communication latency.
- Because WFBP does not reduce the communication overhead, hybrid communication (HybComm) [172] was proposed. Effectively, it combines Parameter Servers (PS) [156] with Sufficient Factor Broadcasting (SFB) [160], choosing the best communication method depending on the sparsity of the parameter tensor. See below for more information about PS (under Centralized Storage) and SFB (under Decentralized Storage).

### 3.6 Discussion

While machine learning and artificial intelligence is a discipline with a long history in computer science, recent advancements in technology have caused certain areas like neural networks to experience unprecedented popularity and impact on novel applications. As with many emerging topics, functionality has been the primary concern and the non-functional aspects have only played a secondary role in the discussion of the technology. As a result, the community has only a preliminary understanding of how distributed machine learning algorithms and systems behave as a workload and which classes of problems have a higher affinity to a certain methodology when considering performance or efficiency.

However, as with similar topics like Big Data Analytics, systems aspects are increasingly becoming more important as the technology matures and consumers are getting more mindful about resource consumption and return of investment. This has caused ML algorithms and systems to be increasingly more co-designed, i.e., adapting algorithms to make better use of systems resources and designing novel systems that support certain classes of algorithms better. We expect this trend to continue and accelerate, eventually leading to a new wave of distributed machine learning systems that are more autonomous in their ability to optimize computation and distribution for given hardware resources. This would significantly lower the burden of adopting distributed machine learning in the same way that popular libraries have democratized machine learning in general by raising the level of abstraction from numerical computing to a simple and approachable templated programming style, or similar to the way that paradigms like MapReduce [44] have made processing of large data sets accessible.

## 4 THE DISTRIBUTED MACHINE LEARNING ECOSYSTEM

The problem of processing a large volume of data on a cluster of machines is not restricted to machine learning but has been studied for a long time in distributed systems and database research. As a result, some practical implementations use general purpose distributed platforms as the foundation for distributed machine learning. Popular frameworks like Apache Spark [169][170] have seized the opportunity of machine learning being an emerging workload and now provide optimized libraries (e.g. MLlib [98]). On the other end of the spectrum, purpose-built machine learning libraries that were originally designed to run on a single machine have started to receive support for execution in a distributed setting. For instance, the popular library Keras [35] received backends to run atop Google's Tensorflow [1] and Microsoft's CNTK [130]. Nvidia extended their machine learning stack with their Collective Communications Library (NCCL) [107] which was originally designed to support multiple GPUs on the same machine but version 2 introduced the ability to run on multiple nodes [76]. The center this ecosystem (Figure 4) is inhabited by systems natively build for distributed machine learning and designed around a specific algorithmic and operational model, e.g. Distributed Ensemble Learning, Parallel Synchronous Stochastic Gradient Descent (SGD) or Parameter Servers. While the majority of these systems are intended to set up and operated by the user and on-premise, there is an increasingly large diversity of machine learning services offered through a cloud delivery model, many centered around established distributed machine learning systems enhanced by a surrounding platform that makes the technology more consumable for data scientists and decision makers.

### 4.1 General Purpose Distributed Computing Frameworks

Distributed systems for processing massive amounts of data largely rely on utilizing a number of commodity servers, each of them with a relatively small storage capacity and computing power, rather than one expensive large server. This strategy has proven more affordable compared to using more expensive specialized hardware, as long as sufficient fault tolerance is built into the software, a concept that Google has pioneered [16] and that has increasingly found traction in the industry. Furthermore, the scale-out model offers a higher aggregate I/O bandwidth compared to using a smaller number of more powerful machines since every node comes with its own I/O subsystem. This can be highly beneficial in data-intensive applications where data ingestion is a significant part of the workload [117].

*4.1.1 Storage.* The storage layer of existing frameworks is commonly based on the *Google File System (GFS)* [55] or comparable implementations. GFS is owned by and used within Google to handle all Big Data storage needs in the company. GFS splits the data that is uploaded to the cluster up into chunks, which are then distributed over the *chunk servers*. The chunks are replicated (the degree of replication is configurable and the default is three-way [55]) in order to protect the data from becoming unavailable in the event of machines failures. The data on the chunk servers can then be accessed by a user through contacting the master, which serves as a name node and provides the locations for every chunk of a file. The GFS architecture was adopted by an open-source framework called Hadoop [104] which was initially developed by Yahoo!, and is now open source and maintained at the Apache Foundation. Its storage layer, named Hadoop File System or HDFS [142], started off as essentially a copy of the GFS design with only minor differences in nomenclature.

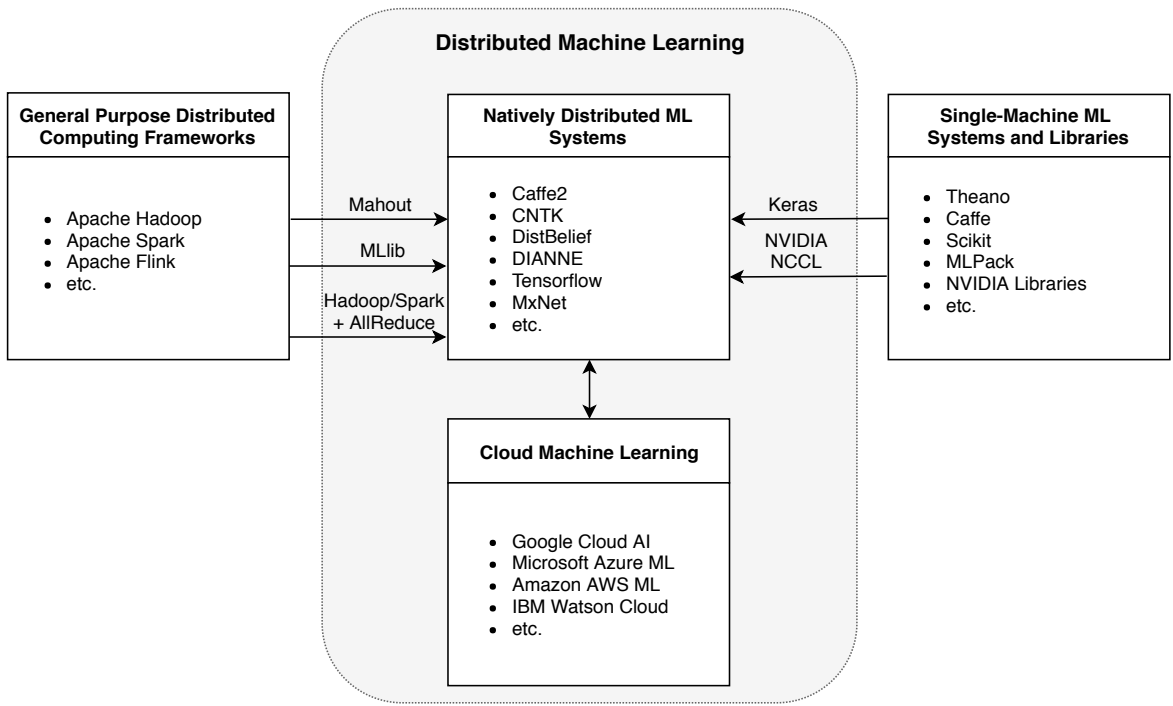

Fig. 4. Distributed Machine Learning Ecosystem. Both general purpose distributed frameworks and single-machine ML systems and libraries are converging towards Distributed Machine Learning. Cloud emerges as a new delivery model for ML.

*4.1.2 Compute.* While the storage architecture has essentially converged to a block-based model, there exist many competing frameworks for scheduling and distributing tasks to compute resources with different features and trade-offs.

*MapReduce.* is a framework (and underlying architecture) for processing data, and was developed by Google [44] in order to process data in a distributed setting. The architecture consists of multiple phases and borrows concepts from functional programming. First, all data is split into tuples (called key-value pairs) during the *map phase*. This is comparable to a mapping of a second-order function to a set in functional programming. The map phase can be executed fully parallel since there are no data dependencies between mapping a function to two different values in the set. Then, during the *shuffle phase*, these tuples are exchanged between nodes and passed on. This is strictly necessary since aggregation generally has data dependencies and it has to be ensured that all tuples belonging to the same key are processed by the same node for correctness. In the subsequent *reduce phase* the aggregation is performed on the tuples to generate a single output value per key. This is similar to a fold operation in functional programming which rolls up a collection using a second-order function that produces a single result value. Fold, however, cannot be parallelized since every fold step depends on the previous step. Shuffling the data and reducing by key is the enabler of parallelism in the reduce phase.

The main benefit of this framework is that the data can be distributed across a large number of machines while tasks of the same phase have not data dependencies and can therefore be executed entirely in parallel. Those same machines can be nodes in a GFS (or similar) storage cluster, so that instead of moving data to the program, the program can be moved to the data for an increase of data locality and better performance. The program is usually several orders of magnitude smaller to transfer over the wire, and is therefore much more efficient to pass around. Furthermore, in compliance with the idea of scale-out, MapReduce implements fault-tolerance in software by monitoring the health of the worker nodes through heartbeat messages and rescheduling tasks that failed to healthy nodes. Typically, the granularity of a task equals the size of a single block in the input data set so that a node failure should only affect a fraction of the overall application and the system is able to recover gracefully. Chu et al. [36] have mapped several machine learning algorithms to the MapReduce framework in order to exploit parallelism for muilticore machines.

The MapReduce architecture is similar to the *Bulk-Synchronous processing (BSP)* paradigm, which preceded it. However, there are some subtle differences. For instance, the MapReduce framework does not allow communication between worker nodes in the map phase. Instead, it only allows cross-communication during the shuffle phase, in between the map and reduce phases [116], for a reduction of synchronization barriers and an increase in parallelism. Goodrich et al. [59] have shown that all BSP programs can be converted into MapReduce programs. Pace [116], in turn, proposed that all MapReduce applications should be modeled as BSP tasks in order to combine the benefits of theoretical correctness of the BSP paradigm with the efficient execution of MapReduce.

MapReduce as a framework is proprietary to Google. The architecture behind it, however, has been recreated in the aforementioned open source Hadoop framework. It leverages HDFS where MapReduce uses GFS, but is similar in its overall architecture. Advanced variants have deliberated themselves from the strict tree topology of MapReduce data flows towards more flexible structures like Forests (Dryad [75]) or generic Directed Acyclic Graphs (DAGs).

*Apache Spark.* MapReduce and Hadoop heavily rely on the distributed file system in every phase of the execution. Even intermediate results are stored on the storage layer, which can be a liability for iterative workloads that need to access the same data repeatedly. Transformations in linear algebra, as they occur in many machine learning algorithms, are typically highly iterative in nature. Furthermore, the paradigm of map and reduce operations is not ideal to support the data flow of iterative tasks since it essentially restricts it to a tree-structure [86] Apache Spark has been developed in response to this challenge. It is capable of executing a directed acyclic graph of transformations (like mappings) and actions (like reductions) fully in memory [138]. Because of its structure, Spark can be significantly faster than MapReduce for more complex workloads. When for example two consecutive map phases are needed, two MapReduce tasks would need to be executed, both of which would need to write all (intermediate) data to disk. Spark, on the other hand, can keep all the data in memory, which saves expensive reads from the disk.

The data structure which Spark was originally designed around is called a *Resilient Distributed Dataset (RDD)*. Such datasets are read-only, and new instances can only be created from data stored on the disk or by transforming existing RDDs [168]. The *Resilient* part comes into play when the data is lost: each RDD is given a lineage graph that shows what transformations have been executed on it. This lineage graph ensures that, if some data is lost, Spark can trace the path the RDD has followed from the lineage graph and recalculate any lost data. It is important that the lineage graph does not contain cycles (i.e. is a Directed Acylic Graph). Otherwise Spark would run into infinite loops and be unable to recreate the RDD. In practice, the need for re-computation as a result of data loss due to node failure can lead to ripple-effects [168]. Spark allows for checkpointing of data to prevent extensive re-computation. Checkpoints have to be explicitly requested and essentially materialize the intermediate state while truncating the RDD lineage graph. Systems like TR-Spark [164] have automated the generation of checkpoints to make Spark able to run on transient resources where interruption of the execution has to be considered the norm.

Apache Spark also includes MLlib, a scalable machine learning library that implements many ML algorithms for classification, regression, decision trees, clustering and topic modeling. It also provides several utilities for building ML workflows, implementing often used feature transformations, hyperparameter tuning, etc. As MLlib uses Spark's APIs, it immediately benefits from the scale-out and failure resilience features of Spark. MLLib relies on the Scala linear algebra package Breeze [64], which in turn utilizes netlib-java [99] for optimization, a bridge for libraries like BLAS [20] and LAPACK [9] which are widely used in high-performance computing.

## 4.2 Natively Distributed Machine Learning Systems

As a result of the rising popularity of Machine Learning in many applications, several domain-specific frameworks have been developed around specific distribution models. In this section, the characteristics of the most popular implementations are summarized.

*4.2.1 Distributed Ensemble Learning.* Many generic frameworks and ML libraries have limited support for distributed training, even though they are fast and effective on a single machine. One way to achieve distribution with these frameworks is through training separate models for subsets of the available data. At prediction time, the outputs of those instances can then be combined through standard ensemble model aggregation [112].

Models that follow this strategy are not dependent on any specific library. They can be orchestrated using existing distribution frameworks (such as MapReduce [44]). The training process involves training individual models on independent machines in parallel. Neither orchestration nor communication are necessary once training has started. Training on $m$ machines with $m$ subsets of the data results in $m$ different models. Each of these can use separate parameters or even algorithms. At prediction time, all trained models can then be run on new data, after which the output of each one is aggregated. This can once again be distributed if needed.

One large drawback is that this method is dependent on proper subdivision of the training data. If large biases are present in the training sets of some of the models, those instances could cause biased output of the ensemble. If the data is divided manually, it is paramount to ensure independence and identical distribution of the data (i.i.d.). If, on the other hand, the dataset is inherently distributed, this is not straightforward to achieve.

There is a large number of existing frameworks available for this method as any Machine Learning framework can be used. Some popular implementations use Tensorflow [1], MXNet [33] and PyTorch [118].

*4.2.2 Parallel Synchronous Stochastic Gradient Descent.* Synchronized parallelism is often the most straightforward to program and reason about. Existing distribution libraries (such as Message Passing Interface (MPI) [62]) can typically be reused for this purpose. Most approaches rely on the AllReduce operation [5] where the compute nodes are arranged in a tree-like topology. Initially, each node calculates a local gradient value, accumulates these with the values received from it's children and sends these up to it's parent (reduce phase). Eventually, the root node obtains the global sum and broadcasts this back down up to the leaf nodes (broadcast phase). Then each node updates its local model with regard to the received global gradient.

*Baidu AllReduce.* uses common high performance computing technology (mainly MPI and its AllReduce operation) to iteratively train SGD models on separate mini-batches of the training data [56]. AllReduce is used to apply each of the workers' gradients onto the last common model state after each operation and then propagate the result of that operation back to each worker. This is an inherently synchronous process, blocking on the result of each workers' training iteration before continuing to the next.

Baidu includes a further optimization from Patarasuk and Yuan [119] in this process, called a Ring AllReduce, to reduce the required amount of communication. By structuring the cluster of machines as a ring (with each node having only 2 neighbors) and cascading the reduction operation, it is possible to utilize all bandwidth optimally. The bottleneck, then, is the highest latency between neighboring nodes.

Baidu claims linear speedup when applying this technique to train deep learning networks. However, it has only been demonstrated on relatively small clusters (5 nodes each, though each node has multiple GPUs that communicate with each other through the same system). The approach lacks fault tolerance by default, as no node in the ring can be missed. This could be counteracted using redundancy (at cost of efficiency). If this is not done, however, the scalability of the method is bounded by the probability of all nodes being available. This probability can be low when using large numbers of commodity machines and networking, which is needed to facilitate Big Data. Baidu's system has been integrated into Tensorflow as an alternative to the built-in Parameter Server-based approach (described below).

*Horovod [132].* takes a very similar approach to that of Baidu: it adds a layer of AllReduce-based MPI training to Tensorflow. One difference is that Horovod uses the NVIDIA Collective Communications Library (NCCL) for increased efficiency when training on (Nvidia) GPUs. This also enables use of multiple GPUs on a single node. Data-parallelizing an existing Tensorflow model is relatively simple since only a few lines of code need to be added, wrapping the default Tensorflow training routine in a distributed AllReduce operation. When benchmarked on Inception v4 [149] and ResNet-101 [68] using 128 GPUs, the average GPU utilization is about 88% compared to about 50% in Tensorflow's Parameter Server approach. However, Horovod lacks fault tolerance (just like in Baidu's approach) and therefore suffers from the same scalability issues [53].

*Caffe2.* (primarily maintained by Facebook) distributes ML through, once again, AllReduce algorithms. It does this by using NCCL between GPUs on a single host, and custom code between hosts based on Facebook's Gloo [47] library to abstract away different interconnects. Facebook uses Ring AllReduce (which offers better bandwidth & parallelism guarantees) but also recursive halving and doubling (a divide-and-conquer approach that offers better latency guarantees). According to their paper, this improves performance in latency-limited situations, such as for small buffer sizes and large server counts. He et al. [68] managed to train ResNet-50 in the span of 1 hour [61] using this approach, achieving linear scaling with the number of GPUs. They achieved 90% efficiency, measured up to 352 GPUs. However, once again no fault-tolerance is present.

*CNTK or The Microsoft Cognitive Toolkit.* offers multiple modes of data-parallel distribution. Many of them use the Ring AllReduce tactic as previously described, making the same trade-off of linear scalability over fault-tolerance. The library offers two innovations:

- **1-bit stochastic gradient descent** (Seide et al. [131]) is an implementation of SGD that quantizes training gradients to a single bit per value. This reduces the number of bits that need to be communicated when doing distributed training by a large constant factor.
- **Block-momentum SGD** (Chen and Huo [31]) divides the training set into m blocks and n splits. Each of the n machines trains a split on each block. Then the gradients calculated for all splits within a block are averaged to obtain the weights for the block. Finally, the block updates are merged into the global model while applying block-level momentum and learning rate.

When benchmarked on a Microsoft speech LSTM, average speedups of 85%+ are achieved for small numbers of GPUs (up to 16), but scalability drops significantly (below 70%) when scaling past that. However, the direct comparison of this number to the other synchronous frameworks' results is questionable, as the dependency structure of an LSTM is significantly different than that of an ordinary DNN due to the introduction of temporal state [140].

*4.2.3 Parallel Asynchronous Stochastic Gradient Descent and Parameter Servers.* Asynchronous approaches tend be more complex to implement and it can be more difficult to trace and debug runtime behavior. However, asynchronism alleviates many problems that occur in clusters with high failure rates or inconsistent performance due to the lack of frequent synchronization barriers.

*DistBelief [43].* is one of the early practical implementations of large-scale distributed ML, and was developed by Google. They encountered the limitations of GPU training and built DistBelief to counteract them. DistBelief supports data- and model-parallel training on tens of thousands of CPU cores (though GPU support was later introduced as well [2]). They reported a speedup of more than 12x when using 81 machines training a huge model with 1.7 billion parameters.

To achieve efficient model-parallelism, DistBelief exploits the structure of neural networks and defines a model as a computation graph where each node implements an operation transforming inputs to outputs. Every machine executes the training of a part of the computation graph's nodes, which can span subsets of multiple layers of the neural network. Communication is only required at those points where a node's output is used as the input of a node trained by another machine.

Partitioning the model across a cluster is transparent and requires no structural modifications. However, the efficiency of a given partitioning is greatly affected by the architecture of the model, and requires careful design. For example, locally connected models lend themselves better for model-parallelism because of limited cross-partition communication. In contrast, fully connected models have more substantial cross-partition dependencies and are therefore harder to efficiently distribute through DistBelief.

To further parallelize model training, data parallelism is applied on top of the model parallelism. A centralized sharded Parameter Server is used to allow each of a set of model replicas (which may be model-parallel internally) to share parameters. DistBelief supports two different methods of data parallelism, both of which are resilient to processing speed variance between model replicas as well as replica failure:

- **Downpour Stochastic Gradient Descent** is an asynchronous alternative to the inherently sequential SGD. Each replica of the model fetches the latest model parameters from the Parameter Server every $n_{fetch}$ steps, updates these parameters in accordance with the model, and pushes the tracked parameter gradients to the Parameter Server every $n_{push}$ steps. The parameters $n_{fetch}$ and $n_{push}$ can be increased to achieve lower communication overhead. Fetching and pushing can happen as a background process, allowing training to continue. Downpour SGD is more resilient to machine failures than SGD, as it allows the training to continue even if some model replicas are off-line. However, the optimization process itself becomes less predictable due to parameters that are out of sync. The authors "found relaxing consistency requirements to be remarkably effective", but offer no theoretical support for this. Tactics that contribute to robustness are the application of adaptive learning rates through AdaGrad [45] and *warm starting* the model through training a single model replica for a while before scaling up to the full number of machines. The authors make note of the absence of stability issues after applying these.
- **Distributed L-BGFS** makes use of an external coordinator process that divides training work between model replicas, as well as some operations on the parameters between the parameter server shards. Training happens through L-BGFS, as is clear from the name.

Each of the shards of the Parameter Server hold a fraction of the parameter space of a model. The model replicas pull the parameters from all shards and each parallelized part of the model only retrieves those parameters that it needs.

Performance improvements are high but the methodology is very expensive in terms of computational complexity. While the best speedup (downpour SGD with AdaGrad) achieved an 80% decrease in training time on ImageNet, this was achieved by using more than 500 machines and more than 1000 CPU cores. It has to be noted that DistBelief did not support distributed GPU training at the time of Dean et al. [43] which could reduce the required resources significantly and is used in fact by almost all other implementations mentioned in this section.

*DIANNE (DIstributed Artificial Neural NEtworks) [39].* is a Java-based distributed deep learning framework using the Torch native backend for executing the necessary computations. It uses a modular OSGi-based distribution framework [155] that allows to execute different components of the deep learning system on different nodes of the infrastructure. Each basic building block of a neural network can be deployed on a specific node, hence enabling model-parallelism. DIANNE also provides basic learner, evaluator and parameter server components that can be scaled and provide a downpour SGD implementation similar to DistBelief.

*Tensorflow [1][2].* is the evolution of DistBelief, developed to replace DistBelief within Google. It borrows the concepts of a computation graph and parameter server from it. It also applies subsequent optimizations to the parameter server model, such as optimizations for training convolutional neural networks Chilimbi et al. [34] and innovations regarding consistency models and fault toleranceLi et al. [92][93]. Unlike DistBelief, TensorFlow was made available as open source software.

TensorFlow represents both model algorithms and state as a dataflow graph, of which the execution can be distributed. This facilitates different parallelization schemes that can take e.g. state locality into account. The level of abstraction of the dataflow graph is mathematical operations on tensors (i.e. $n$-dimensional matrices). This in contrast to DistBelief, which abstracts at the level of individual layers. Consequently, defining a new type of neural network layer in Tensorflow requires no custom code - it can be represented as a subgraph of a larger model, composed of fundamental math operations. A Tensorflow model is first defined as a symbolic dataflow graph. Once this graph has been constructed, it is optimized and then executed on the available hardware. This execution model allows Tensorflow to tailor its operations towards the types of devices available to it. When working with, e.g., GPUs or TPUs (Tensor Processing Units [80]), Tensorflow can take into account the asynchronicity and intolerance or sensitivity to branching that is inherent to these devices, without requiring any changes to the model itself.

Shi and Chu [139] shows Tensorflow achieving about 50% efficiency on 4-node, InfiniBand-connected cluster training of ResNet-50He et al. [68], and about 75% efficiency on GoogleNet [148], showing that the communication overhead plays an important role, and also depends on architecture of the neural network to optimize.

*MXNet [33].* uses a strategy very similar to that of Tensorflow: models are represented as dataflow graphs, which are executed on hardware that is abstracted away and coordinated by using a parameter server. However, MXNet also supports the imperative definition of dataflow graphs as operations on n-dimensional arrays, which simplifies the implementation of certain kinds of networks.

MXNet's Parameter Server, KVStore, is implemented on top of a traditional key-value store. The KVStore supports pushing key-value pairs from a device to the store, as well as pulling the current value of a key from the store. There is support for user-defined update logic that is executed when a new value is pushed. The KVStore can also enforce different consistency models (currently limited to sequential and eventually consistent execution). It is a two-tier system: updates by multiple threads and GPUs are merged on the local machine before they're pushed to the full cluster. The KVStore abstraction theoretically enables the implementation of (stale-)synchronicity, although only an asynchronous implementation is present at the time of writing.

On a small cluster of 10 machines equipped with a GPU, MXNet achieves almost linear speedup compared to a single machine when training GoogleNet [148] with more than 10 passes over the data [33].

*DMTK or the Distributed Machine Learning Toolkit [103].* from Microsoft includes a Parameter Server called *Multiverso*. This can be used together with CNTK to enable Asynchronous SGD instead of the default Allreduce-based distribution in CNTK.

### 4.2.4 Parallel Stale-synchronous Stochastic Gradient Descent.

*Petuum [161].* aims to provide a generic platform for any type of machine learning (as long as it is iteratively convergent) on big data and big models (hundreds of billions of parameters). It supports data- and model-parallelism. The Petuum approach exploits ML's error tolerance, dynamic structural dependencies, and non-uniform convergence in order to achieve good scalability on large datasets and models. This is in contrast to for example Spark, which focuses on fault tolerance and recovery. The platform uses stale synchronicity to exploit inherent tolerance of machine learning against errors since a minor amount of staleness will only have minor effects on convergence.

Dynamic scheduling policies are employed to exploit dynamic structural dependencies which helps minimize parallelization error and synchronization cost. Finally, unconverged parameter prioritization takes advantage of non-uniform convergence by reducing computational cost on parameters that are already near optimal.

Petuum uses the Parameter Server paradigm to keep track of the parameters of the model being trained. The Parameter Server is also responsible for maintaining the staleness guarantees. In addition, it exposes a scheduler that lets the model developer control the ordering of parallelized model updates.

When developing a model using Petuum, developers have to implement a method named *push*, which is responsible for each of the parallelized model training operations. Its implementation should pull the model state from the parameter server, run a training iteration, and push a gradient to the parameter server. Petuum by default manages the scheduling aspect and the parameter merging logic automatically, so that data-parallel models don't require any additional operations. However, if model-parallelism is desired, the schedule method (which tells each of the parallel workers what parameters they need to train) and the pull method (which defines the aggregation logic for each of the generated parameter gradients) need to be implemented as well.

Petuum provides an abstraction layer that also allows it to run on systems using YARN (the Hadoop job scheduler) and HDFS (the Hadoop file system), which simplifies compatibility with pre-existing clusters.

*4.2.5 Parallel Hybrid-synchronous SGD.* Both synchronous and asynchronous approaches have some significant drawbacks, as is explored by Chen et al. [30]. A few frameworks attempt to find a middle ground instead that combines some of the best properties of each model of parallelism and diminishes some of the drawbacks.

*MXNet-MPI [96].* takes an approach to distributed ML (using a modified version of MXNet as a proof of concept) that combines some of the best aspects of both asynchronous (Parameter Server) and synchronous (MPI) implementations. The idea here is to use the same architecture as described in the MXNet section. Instead of having single workers communicate with the parameter server, however, those workers are clustered together into groups that internally apply synchronous SGD over MPI with AllReduce. This has the benefit of easy linear scalability of the synchronous MPI approach and fault tolerance of the asynchronous Parameter Server approach.

## 4.3 Machine Learning in the Cloud

Several cloud operators have added machine learning as a service to their cloud offerings. Most providers offer multiple options of executing machine learning tasks in their clouds, ranging from IaaS-level services (VM instances with pre-packaged ML software) to SaaS-level solutions (Machine Learning as a Service). Much of the technology offered are standard distributed machine learning systems and libraries. Among other things, Google's Cloud Machine Learning Engine offers support for TensorFlow and even provides TPU instances [60]. Microsoft Azure Machine Learning allows model deployment through Azure Kubernetes, through a batch service, or by using CNTK VMs [102]. As a competitor to Google's TPUs, Azure supports accelerating ML applications through FPGAs [115]. Amazon AWS has introduced SageMaker, a hosted service for building and training machine learning models in the cloud. The service includes support for TensorFlow, MXNet, and Spark [7]. IBM has bundled their cloud machine learning offerings under the Watson brand [74]. Services include Jupyter notebooks, Tensorflow, and Keras. The cloud-based delivery model is becoming more important as it reduces the burden of entry into designing smart applications that facilitate machine learning techniques. However, the cloud is not only a consumer of distributed machine learning technology but is also fueling the development of new systems and approaches back to the ecosystem in order to handle the large scale of the deployments.

## 5 CONCLUSIONS AND CURRENT CHALLENGES

Distributed Machine Learning is a thriving ecosystem with a variety of solutions that differ in architecture, algorithms, performance, and efficiency. Some fundamental challenges had to be overcome to make distributed machine learning viable in the first place, such as finding mechanisms to efficiently parallelize the processing of data while combining the outcome into a single coherent model. Now that there are industry-grade systems available and in view of the ever growing appetite for tackling more complex problems with machine learning, distributed machine learning is increasingly becoming the norm and single-machine solutions the exception, similar to how data processing in general had developed in the past decade. There are, however, still many open challenges that are crucial to the long-term success of distributed machine learning.

### 5.1 Performance

A trade-off that is seen frequently is the reduction of wall-clock time at the expense of total aggregate processing time (i.e. decreased efficiency) by adding additional resources. When compute resources are affordable enough, many real-world use cases of machine learning benefit most from being trained rapidly. The fact that this often implies a large increase in total compute resources and the associated energy consumption, is not considered important as long as a model saves more money than it costs to train. A good example of this is found in Dean et al. [43], where wall clock time speedup factors are achieved by increasing the number of machines quadratically or worse. It still delivered Google competitive advantage for years. Distributed use of GPUs, as in Tensorflow, has better properties, but often still exhibits efficiency below 75%. These performance concerns are much less severe in the context of synchronous SGD-based frameworks, which often do achieve linear speedups in benchmarks. However, most of these benchmarks test at most a few hundred machines, whereas the scale at which e.g. DistBelief is demonstrated, can be two orders of magnitude larger. The research community could clearly benefit from more independent studies that report on the performance and scalability of these systems for larger and more realistic applications, and that could provide valuable insights to guide research into workload optimization and system architecture.

### 5.2 Fault Tolerance

Synchronous AllReduce-based approaches seem to scale significantly better than the parameter server approach (up to a certain cluster size), but suffer from a lack of fault-tolerance: failure of a single machine blocks the entire training process. At smaller scales, this might still be a manageable problem. However, past a certain number of nodes the probability of any node being unavailable becomes high enough to result in near-continuous stalling. Common implementations of these HPC-inspired patterns, such as MPI and NCCL, lack fault-tolerance completely. Although there are efforts to counteract some of this, production-ready solutions are lacking. Some of the described implementations allow for checkpointing to counteract this, but significant effort is necessary to enable true fault-tolerance, as is described in Amatya et al. [6]. It is also possible to reduce the probability of failure for each individual node, but this requires very specific hardware that is expensive and not generally available in commodity scale-out data centers or in the cloud. Asynchronous implementations do not suffer from this problem as much. They are designed to explicitly tolerate straggling [41] (slow-running) and failing nodes, with only minimal impact on training performance. The question for ML operators, then, is whether they prefer performance or fault tolerance, and whether they are constrained by either one. Hybrid approaches even offer a way to customize these characteristics, although they are not frequently found in use yet. It would be interesting to see whether an even better approach exists, or whether there is an efficient way to implement fault-tolerant AllReduce.

### 5.3 Privacy

There are scenarios in which it is beneficial or even mandatory to isolate different subsets of the training data from each other [79]. The furthest extent of this is when a model needs to be trained on datasets that each live on different machines or clusters, and may under no circumstance be co-located or even moved. Peer-to-peer topologies like Gossip Learning [113] fully embrace this principle.

Another approach to training models in a privacy-sensitive context is the use of a distributed ensemble model. This allows perfect separation of the training data subsets, with the drawback that a method needs to be found that properly balances each trained model's output for an unbiased result.

Parameter server-based systems can be useful in the context of privacy, as the training of a model can be separated from the training result. Abadi et al. [3] discuss several algorithms that are able to train models efficiently while maintaining differential privacy. These parameter server-based systems assume that no sensitive properties of the underlying data leak into the model itself, which turns out to be difficult in practice. Recently, Bagdasaryan et al. [12] showed that it is possible for attackers to implement a back-door into the joint model.

Federated learning systems can be deployed where multiple parties jointly learn an accurate deep neural network while keeping the data itself local and confidential. Privacy of the respective data was believed to be preserved by applying differential privacy, as shown by Shokri and Shmatikov [141] and McMahan et al. [97]. However, Hitaj et al. [71] devised an attack based on GANs, showing that record-level differential privacy is generally ineffective in federated learning systems.

Additionally, it is possible to introduce statistical noise into each subset of the training data, with the intention of rendering its sensitive characteristics unidentifiable to other parties. Balcan et al. [14] touches on this subject, but makes it clear that the resulting privacy in this scenario is dependent on the amount of statistical queries required to learn the dataset. This puts an upper bound on usefulness of the model itself.

For a more in-depth discussion on privacy in distributed deep learning, we refer to Vepakomma et al. [154]. In conclusion, while theoretical results exist, current frameworks do not offer much support for even basic forms of privacy. It could be interesting to investigate fundamental approaches to facilitate distributed privacy, which could then be integrated into the currently popular frameworks.

### 5.4 Portability

With the proliferation of machine learning, in particular deep learning, a myriad of different libraries and frameworks for creating and training neural networks is established. However, once trained, one is often stuck to the framework at hand to deploy the model in production, as they all use a custom format to store the results. For example, Tensorflow [2] uses a SavedModel directory, which includes a protocol buffer defining the whole computation graph. Caffe [78] also uses a binary protocol buffer for storing saved models, but with a custom schema. Theano [18] uses pickle to serialize models represented by Python objects, and PyTorch [118] has a built-in save method that serializes to a custom ASCII or binary format.

Portability also becomes increasingly important with respect to the hardware platform on which one wants to deploy. Although the x86_64 and ARM processor architectures are mainstream to execute applications in the server and mobile devices market respectively, we witness a shift towards using GPU hardware for efficiently executing neural network models [109]. As machine learning models become more widespread, we also see more and more development of custom ASICs such as TPUs [129] in Google Cloud or dedicated neural network hardware in the latest iPhone [11]. This diversification makes it more difficult to make sure that your trained model can run on any of these hardware platforms.

A first step towards portability is the rise of a couple of framework independent specifications to define machine learning models and computation graphs. The Open Neural Network Exchange (ONNX) format defines a protocol buffer schema that defines an extensible computation graph model, as well as definitions for standard operators and data types. Currently, ONNX is supported out of the box by frameworks such as Caffe, PyTorch, CNTK and MXNet and converters exist, e.g., for TensorFlow. Similar efforts for a common model format specification are driven by Apple with their Core ML format [10] and the Khronos Group with the Neural Network Exchange Format [151].